\documentclass[10pt,twocolumn,letterpaper]{article}

\usepackage[accept]{cvpr}      
\definecolor{cvprblue}{rgb}{0.21,0.49,0.74}
\usepackage[pagebackref,breaklinks,colorlinks,allcolors=cvprblue]{hyperref}
\usepackage{colortbl}
\usepackage{multirow} 
\usepackage{dsfont}
\usepackage[accsupp]{axessibility}

\def\paperID{45590} 
\def\confName{CVPR}
\def\confYear{2026}

\title{From Horizontal to Rotated: Cross-View Object Geo-Localization with Orientation Awareness} 

\author{Chenlin Fu, Ao Gong, Yingying Zhu\thanks{Corresponding author.}\\
College of Computer Science and Software Engineering, Shenzhen University, China\\
{\tt\small fuchenlin2021@email.szu.edu.cn, 2410103027@mails.szu.edu.cn, zhuyy@szu.edu.cn}
}

\begin{document}
\maketitle
\begin{abstract}

Cross-View object geo-localization (CVOGL) aims to precisely determine the geographic coordinates of a query object from a ground or drone perspective by referencing a satellite map. Segmentation-based approaches offer high precision but require prohibitively expensive pixel-level annotations, whereas more economical detection-based methods suffer from lower accuracy. This performance disparity in detection is primarily caused by two factors: the poor geometric fit of Horizontal Bounding Boxes (HBoxes) for oriented objects and the degradation in precision due to feature map scaling. Motivated by these, we propose leveraging Rotated Bounding Boxes (RBoxes) as a natural extension of the detection-based paradigm. RBoxes provide a much tighter geometric fit to oriented objects. Building on this, we introduce OSGeo, a novel geo-localization framework, meticulously designed with a multi-scale perception module and an orientation-sensitive head to accurately regress RBoxes. To support this scheme, we also construct and release CVOGL-R, the first dataset with precise RBox annotations for CVOGL. Extensive experiments demonstrate that our OSGeo achieves state-of-the-art performance, consistently matching or even surpassing the accuracy of leading segmentation-based methods but with an annotation cost that is over an order of magnitude lower.


\end{abstract}    
\section{Introduction}
\label{sec:intro}

\begin{figure}[t]
\centering
\includegraphics[width=0.45\textwidth]{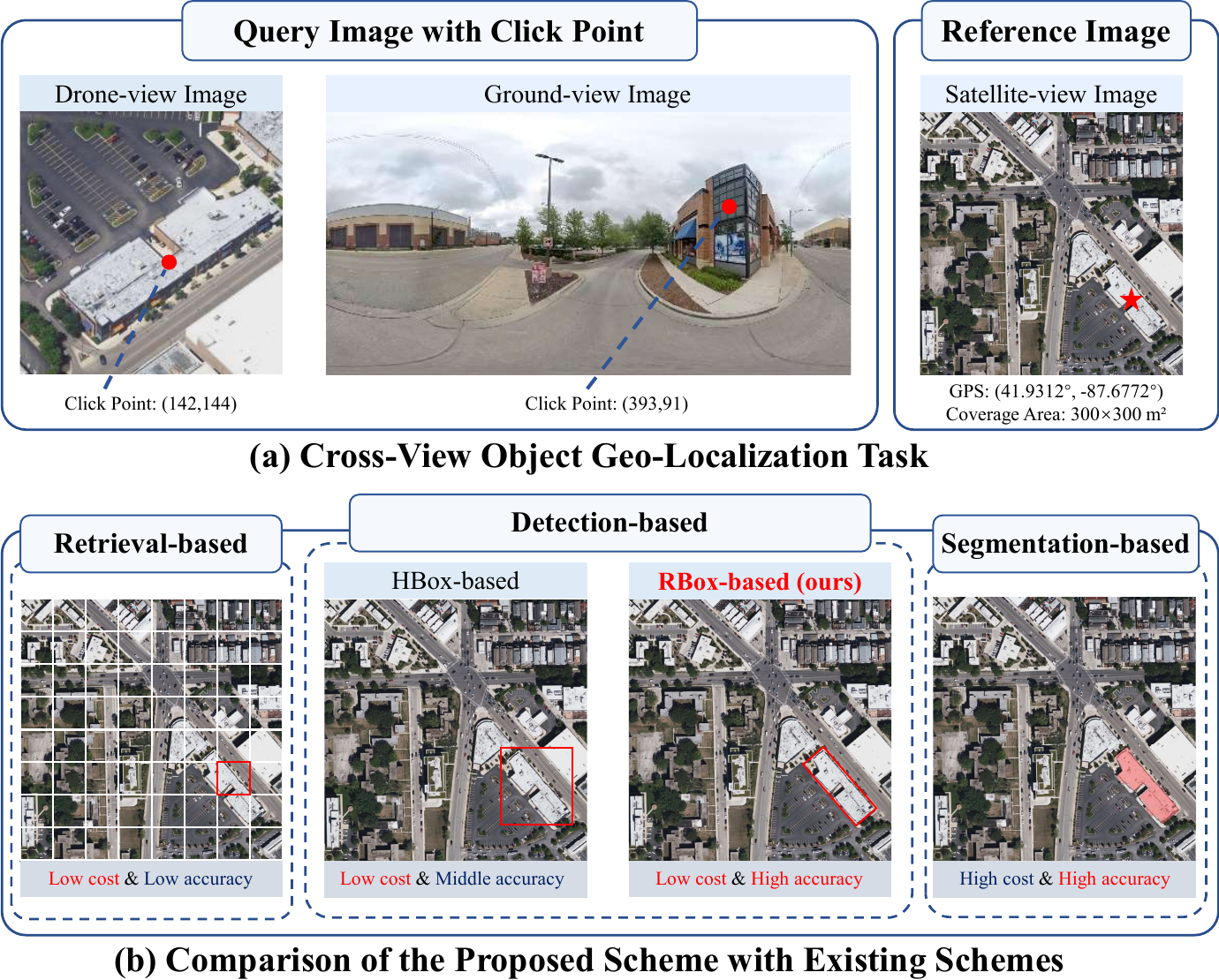}
\caption{Illustration of the CVOGL task and a comparison of solution paradigms. \textbf{(a)} The Cross-View Object Geo-Localization (CVOGL) Task: A query object, indicated by a click point (red dot) in a drone or ground-view image, is to be precisely localized on a georeferenced satellite image. \textbf{(b)} Comparison of Localization Granularity: We compare four paradigms. \textbf{Retrieval-based:} Matches to a coarse grid cell. \textbf{HBox-based (Detection):} Localizes with a loose, axis-aligned Horizontal Bounding Box (HBox). \textbf{RBox-based (Ours):} Localizes with a tight, oriented Rotated Bounding Box (RBox). \textbf{Segmentation-based:} Localizes with a pixel-perfect (but costly) mask.}
\label{fig:intro}
\end{figure}

Cross-View Object Geo-Localization (CVOGL) is the task of pinpointing the geographic location of a query object from a ground or drone query image, using a georeferenced satellite image as a global map (see \cref{fig:intro}(a)). This technique effectively aligns the local perspective of the query with the global overview from the satellite, thereby providing a clear visual context of the query object relative to its surrounding environment. Its significance is particularly pronounced in scenarios where GPS signals are obstructed or perturbed by noise, where it offers an alternative means for geo-localization. Consequently, CVOGL holds substantial value in domains such as smart city management, disaster monitoring, and autonomous navigation \cite{2024florence, 2024good, buildings, smartcities, multimodal}.

As illustrated in \cref{fig:intro}(b), existing approaches to CVOGL have diverged into three main paradigms: retrieval-based, detection-based, and segmentation-based methods. The earliest retrieval-based approaches \cite{jiansuoCVM-Net, jiansuoL2LTR, jiansuoRK-Net, jiansuoSAFA, jiansuosample4geo, jiansuoTransgeo} localize query objects by matching them to a coarse grid of satellite patches. While computationally efficient, their localization precision is inherently limited by the grid granularity. Subsequent detection-based methods \cite{jianceDetgeo, jianceHALO, jianceVAgeo} refined this by introducing Horizontal Bounding Box (HBox) representations, improving precision with only modest annotation costs. More recently, the segmentation-based TROGeo \cite{zhang2025breaking} has achieved the highest localization accuracy by producing pixel-level segmentation, but this comes at a prohibitive annotation cost approximately 15 times \cite{cost2023weakly,coco,maskfree2023} higher than bounding boxes, creating a significant barrier to scalability.

This sharp trade-off between precision and cost motivates a critical re-examination of the task's fundamental requirements. While segmentation-based TROGeo \cite{zhang2025breaking} achieves the highest precision, this accuracy comes at a prohibitive annotation cost. We argue that this cost is largely disproportionate to the core objective of CVOGL. Crucially, the primary goal is to determine an object's geographic coordinates typically its center point, rather than the precise location of every pixel on its silhouette. The substantial effort and expense required for pixel-level supervision can therefore be considered a form of "overkill" relative to this pragmatic goal. This precision-cost dilemma, therefore, necessitates a new solution that can retain high accuracy without the prohibitive expense of mask-level supervision.

To devise a solution that retains high accuracy while reducing annotation cost, it is crucial to analyze the sources of this performance disparity. The superior performance of the segmentation-based method is largely attributed to its pixel-level supervision, which grants it distinct advantages in both \textbf{precise boundary delineation} and \textbf{faithful scale representation}. Conversely, while more cost-efficient, existing detection-based approaches suffer from two fundamental limitations. \textbf{The first and critical factor} is the poor geometric fit of HBox for oriented objects. As depicted in \cref{fig:intro}(b), HBox, designed for axis-aligned shapes, inherently fails to encase oriented objects tightly. This mismatch introduces substantial background noise, which degrades network optimization and hinders accurate positioning. \textbf{The second factor} stems from the inherent architectural limitations of detection-based models, where the final location is regressed from a single down-sampled feature map, leading to spatial quantization that limits precision.

Motivated by these insights, we propose a paradigm shift that directly confronts both limitations. First, to solve the geometric inadequacy of HBoxes, we advocate for leveraging Rotated Bounding Boxes (RBoxes). As illustrated in \cref{fig:intro}(b), RBoxes provide a much tighter geometric fit to oriented objects, significantly reducing background noise and enabling a more faithful representation. Building on this, we introduce OSGeo, a novel Orientation-Sensitive Geo-localization framework. OSGeo is specifically engineered with a multi-scale perception module to address scale variance and an orientation-sensitive head designed for superior geometric fitting. To facilitate this new RBox-based paradigm, we also construct and release CVOGL-R, the first benchmark dataset meticulously re-annotated with precise RBoxes. This holistic approach allows OSGeo to achieve precision that is not only comparable to but often surpasses that of leading segmentation-based methods, thereby resolving the critical trade-off between accuracy and annotation cost.

Our contributions can be summarized as follows:

\begin{itemize}
    \item  We introduce and validate a new RBox-based paradigm for CVOGL, which resolves the critical precision-cost dilemma by proposing a "sweet spot" that avoids the geometric inadequacy of HBoxes and the "overkill" of segmentation's high annotation cost. To facilitate and benchmark this new paradigm, we also construct and release CVOGL-R, the first benchmark dataset meticulously re-annotated with precise RBoxes.
    \item We propose OSGeo, a novel framework explicitly engineered to realize the RBox paradigm's potential. Its components, including a multi-scale perception module and an orientation-sensitive head, are designed to directly address the key limitations (feature map scaling and geometric fit of previous detection-based methods.
    \item Through extensive experiments, our OSGeo framework achieves state-of-the-art performance. Crucially, our OSGeo achieves localization precision that is comparable to, and often surpasses, leading segmentation-based method, but with a dramatically lower annotation cost.
\end{itemize}



\section{Related Work}
\label{sec:related}
\subsection{Cross-View Object Geo-Localization}

\noindent
\textbf{Retrieval-Based.} The initial works for cross-view object geo-localization treat the task as a fine-grained image retrieval problem \cite{jiansuoCVM-Net, cvigl2013, cvigl2015wide}. These methods involve dividing the satellite map into a grid of uniform patches to find the one with the highest feature similarity to the query object \cite{jiansuoL2LTR, jiansuoRK-Net, jiansuoSAFA, jiansuosample4geo, jiansuoTransgeo}. However, its precision is fundamentally restricted by the fixed grid structure. The coarse granularity of these predefined patches limits localization accuracy and fails to accommodate significant variations in object size.

\noindent
\textbf{Segmentation-Based.} In pursuit of higher precision, TROGeo \cite{zhang2025breaking} proposes a cross-view object segmentation (CVOS) scheme to reformulate cross-view object geo-localization, which uses segmentation masks to define objects for fine-grained object geo-localization. Despite the high precision of this paradigm, it faces a significant bottleneck related to data acquisition.

\noindent
\textbf{Detection-Based.} Detection-based methods \cite{jianceDetgeo,jianceHALO,jianceVAgeo} offer a more pragmatic alternative by defining object locations with horizontal bounding boxes (HBox). These methods exhibit greater flexibility, as they can be augmented with class-agnostic segmentation models like SAM \cite{sam} to generate fine-grained masks without requiring dedicated segmentation datasets. We unlock the potential of detection-based approaches by incorporating object orientation and multi-scale features, achieving performance comparable to segmentation-based methods.


\subsection{Oriented Object Detection}
Oriented object detection constitutes a fine-grained perception task characterized by diverse methodological paradigms. Initial efforts focus on architectural design, progressing from anchor-based \cite{retina} and two-stage methods \cite{OrientedR-CNN, roitrans, redet} to anchor-free frameworks \cite{fcos}. Subsequently, a significant body of work addresses the fundamental challenge of angle periodicity. Solutions include designing specialized loss functions \cite{RSDET}, developing boundary-free angle-coding schemes \cite{CSL, DCL, PSC}, and modeling the rotated bounding box (RBox) with Gaussian distributions \cite{GWD, KLD}. In parallel, an alternative line of research utilizes representative point sets to define object boundaries, as exemplified by RepPoints \cite{reppoints, G-rep, o-rep}.

We draw inspiration from these methods, utilizing its precise localization capabilities to address the geometric inadequacy of HBoxes in cross-view object geo-localization.

\section{Methodology}

\subsection{Why Rotated Bounding Boxes}
\label{sec:pre}
\begin{figure}[t]
\centering
\includegraphics[width=0.82\linewidth]{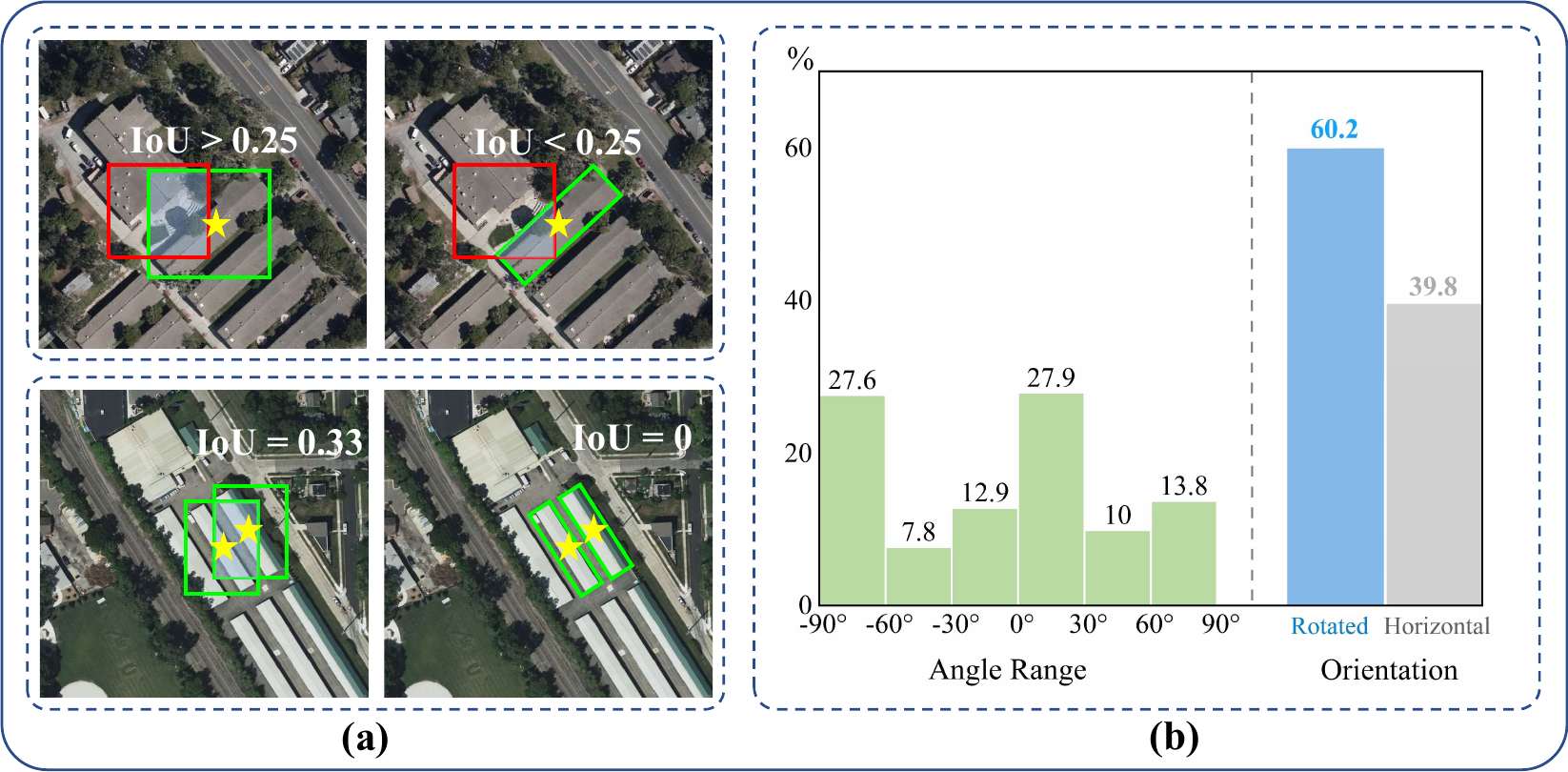}
\caption{Illustration of the necessity of RBoxes. The star denotes the object center. The \textcolor{red}{\emph{red}} rectangle represents the predicted box, while the \textcolor{green}{\emph{green}} rectangle denotes the ground truth box.}
\label{fig:h-r}
\end{figure}

To motivate our approach, we first highlight three critical failures of the standard Horizontal Bounding Box (HBox) representation in CVOGL. \textbf{First}, as illustrated in the top row of \cref{fig:h-r}(a), a predicted bounding box can achieve a high IoU with the GT HBox while failing to localize the actual rotated object, resulting in a near-zero IoU with its corresponding rotated RBox. This discrepancy has a profound quantitative impact. When re-evaluated using the more precise RBox criterion, the performance of DetGeo \cite{jianceDetgeo} on the Drone→Satellite test set plummets by a substantial \textbf{14.13\%} in Acc@50, as detailed in \cref{tab:comparison_cvogl-R}. \textbf{Second}, HBoxes introduce severe ambiguity in dense scenes. The example in the bottom row of \cref{fig:h-r}(a) shows adjacent objects with a high HBox IoU of \textbf{0.33}, complicating instance differentiation. In contrast, their RBoxes maintain a \textbf{near-zero} IoU, ensuring clear object separation. \textbf{Finally}, orientation is not an edge case but a ubiquitous feature in satellite imagery. Our statistical analysis reveals that \textbf{60.2\%} of instances in the benchmark dataset are rotated, establishing orientation modeling as an essential component for accurate geo-localization. 



\begin{figure*}[ht]
\centering
\includegraphics[width=0.8\textwidth]{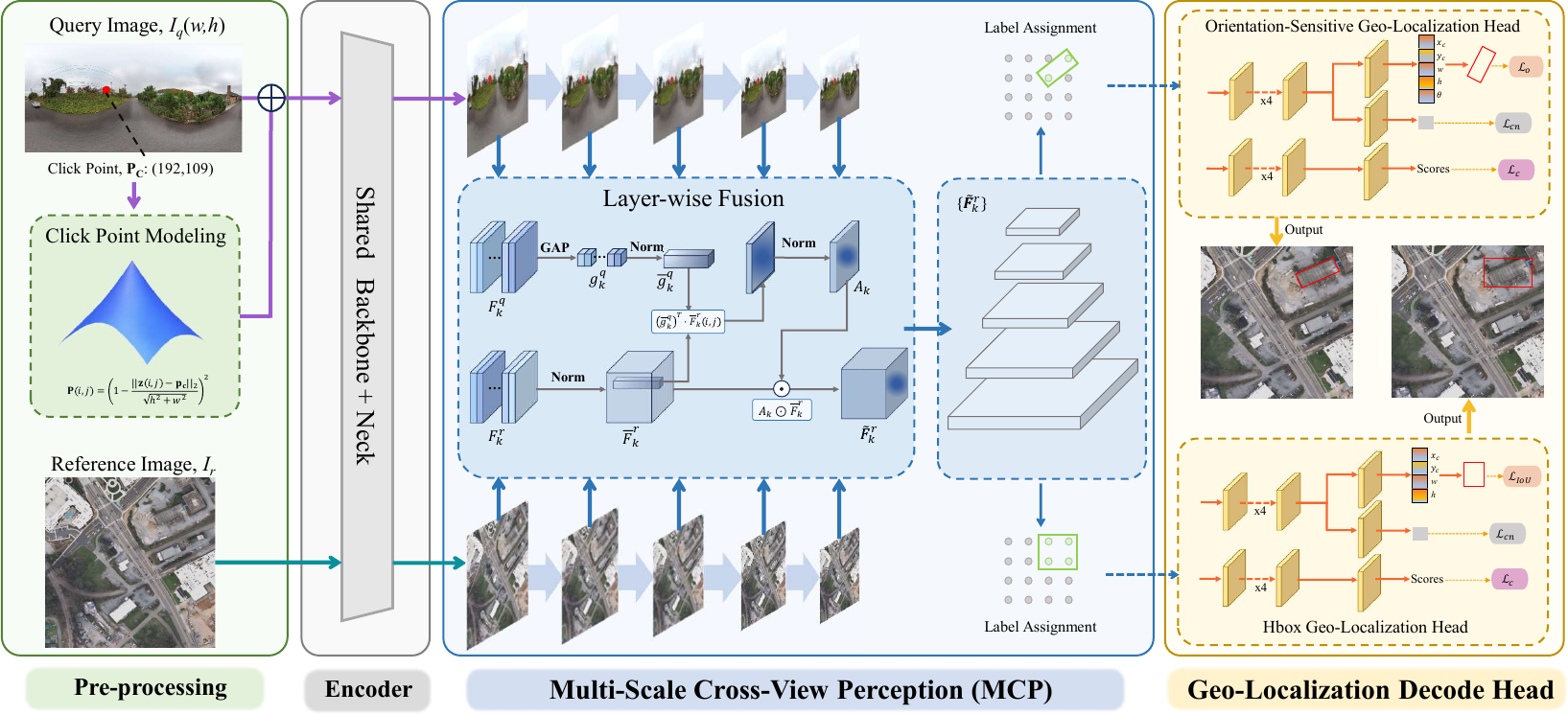} 
\caption{Overall architecture of our proposed OSGeo. The architecture consists of a Pre-processing for input preparation and click point modeling, an Encoder for feature extraction, an MCP for multi-scale feature fusion, and a Geo-Localization Decode Head for prediction.}
\label{fig:pip}
\end{figure*}

\subsection{OSGeo}
\textbf{Overview.} In this section, we introduce our OSGeo. As illustrated in \cref{fig:pip}, OSGeo reframes the conventional detection-based approach by adopting a modular architecture. This architecture comprises four main stages: Pre-processing, Encoder, the Multi-Scale Cross-View Perception (MCP) module, and a Geo-Localization Decode Head. The MCP module is specifically designed to mitigate precision loss from feature down-sampling by fusing features across multiple scales. The Geo-Localization Decode Head, in turn, integrates two specialized sub-heads: an Orientation-Sensitive Geo-Localization Head and an HBox Geo-Localization Head. This dual-head design enables the framework to switch freely between horizontal and rotational regression. To optimize the latter, we also propose OS-Loss, an orientation-sensitive loss function carefully tailored for this purpose. Furthermore, the decoded bounding boxes can serve as effective prompts for the SAM \cite{sam}, facilitating finer-grained segmentation.

\subsubsection{Pre-Processing and Encoder}
\textbf{Click Point Modeling.} To construct an explicit spatial representation of the click point $\mathbf{p}_c$, we introduce a click representation map, denoted as $\mathbf{P}$. Let the coordinate of the single click point be $\mathbf{p}_c$. For a query image $I_q$ with a height $h$ and a width $w$, we define this map over a discrete grid. The value of the map for each grid element is computed as:
\begin{equation}
\mathbf{P}(i, j) = \left( 1 - \frac{\|\mathbf{z}(i, j) - \mathbf{p}_c\|_2}{\sqrt{h^2 + w^2}} \right)^2
\label{eq:p}
\end{equation}
where the indices $(i, j)$ span the entire image dimensions, with $0 \le i < h$ and $0 \le j < w$, $\mathbf{z}(i, j)$ represents the vector for the coordinate $(i, j)$, a three-dimensional visualization of this map is presented in \cref{fig:pip}. Subsequently, the generated map $\mathbf{P}$ is then channel-wise concatenated with the query image $I_q$, denoted as $I^p_q$.

\noindent
\textbf{Shared Backbone and Neck.} The combined query tensor $I^p_q$, produced by the click point modeling, and the reference image $I_r$ are concurrently processed through a shared feature extraction encoder. This encoder consists of a ResNet-50 \cite{resnet50} backbone and a Feature Pyramid Network (FPN) \cite{FPN} neck. Critically, the weights of this aforementioned encoder are shared for processing both the query and reference inputs. This weight-sharing mechanism is essential for projecting the cross-view image pair into a common embedding space. The FPN then generates multi-scale feature maps, $\{F^q_k\}$ and $\{F^r_k\}$, for the query and reference inputs, respectively. The index $k$ signifies the pyramid level, enabling the model to represent objects at various scales effectively. Following the standard FPN architecture, we extract features from five levels, where $k \in \{3, 4, 5, 6, 7\}$. Levels 3 through 5 originate from the ResNet backbone outputs with top-down connections, while levels 6 and 7 are derived by applying strided convolutions from the preceding level. The resulting sets of multi-scale features, $\{F^q_k\}$ and $\{F^r_k\}$, are then passed directly to the MCP.

\subsubsection{Multi-Scale Cross-View Perception Module}
A primary source of imprecision in existing detection-based methods is their reliance on heavily down-sampled feature maps for coordinate regression. This process of spatial quantization invariably discards fine-grained spatial information crucial for pinpoint localization. To address this limitation, we design the Multi-Scale Cross-View Perception (MCP) module. The core function of the MCP module is to create a rich, multi-resolution feature representation by establishing interactions between the query and reference views at different levels of the feature hierarchy. This interaction is performed in a layer-by-layer manner across the corresponding multi-scale feature maps $\{F^q_k\}$ and $\{F^r_k\}$.

At each feature level $k \in \{3, ..., 7\}$, the MCP refines the reference features $F^r_k$ using guidance from the corresponding query features $F^q_k$. The entire query feature map $F^q_k \in \mathbb{R}^{D_k \times H_k \times W_k}$ is first aggregated into a single global context vector. This is achieved by applying global average pooling (GAP) across its spatial dimensions. This global query vector $\mathbf{g}^q_k \in \mathbb{R}^{D_k}$ encapsulates the holistic representation of the query target at scale $k$.

To measure the similarity between the global query context and every local region in the reference view, we compute an attention score map. First, both the global query vector $\mathbf{g}^q_k$ and each local feature vector $F^r_k(i,j) \in \mathbb{R}^{D_k}$ from the reference map are L2-normalized. The score $s_k(i,j)$ at each spatial location $(i,j)$ is then computed as the dot product between the normalized vectors:
\begin{equation}
s_k(i,j) = \frac{(\mathbf{g}^q_k)^T}{||\mathbf{g}^q_k||_2} \cdot \frac{F^r_k(i,j)}{||F^r_k(i,j)||_2}
\label{eq:score}
\end{equation}
where $0 \le i < H_k$ and $0 \le j < W_k$.

The raw scores $\{s_k(i,j)\}$ are then normalized to a $[0, 1]$ range to form a stable attention map $A_k$. This is accomplished using min-max scaling applied across all spatial locations for each sample in the batch:
\begin{equation}
A_k(i,j) = \frac{s_k(i,j) - \min(s_k)}{\max(s_k) - \min(s_k) + \epsilon}
\label{eq:attn}
\end{equation}
where $\min(s_k)$ and $\max(s_k)$ represent the minimum and maximum score values across the entire spatial map for level $k$, and $\epsilon$ is a small constant to prevent division by zero. This step yields the final attention map $A_k \in \mathbb{R}^{1 \times H_k \times W_k}$.

Finally, the computed attention map $A_k$ is used to modulate the normalized original reference feature map $\overline{F}^r_k$. The refined reference feature map, denoted as $\tilde{F}^r_k$, is obtained through element-wise multiplication:
\begin{equation}
\tilde{F}^r_k = A_k \odot \overline{F}^r_k
\label{eq:context}
\end{equation}
where $\odot$ denotes the element-wise product, the resulting $\tilde{F}^r_k$ highlights the regions in the reference view that are most relevant to the global context of the query target.


This attention mechanism is performed independently and in parallel for each feature level $k \in \{3, ..., 7\}$. The final output from the MCP module is the complete set of refined, multi-scale reference features $\{\tilde{F}^r_k\}$, which are then passed to the subsequent Geo-localization Decode Head.

\subsubsection{Geo-Localization Decode Head}
The Geo-Localization Decode Head processes the refined multi-scale features $\{\tilde{F}^r_k\}$ to produce the final prediction. 

\noindent
\textbf{Orientation-Sensitive Geo-Localization Head.} Inspired by FCOS \cite{fcos}, for each feature level $k$ with stride $s_k$, a location $(i, j)$ on the feature map $\tilde{F}^r_k$ is considered a positive sample if its corresponding location $(i_m, j_m)$ on the input image falls within any GT RBox. The mapping from feature map coordinates to image coordinates is defined as:
\begin{equation}
(i_m, j_m) = \left( \left\lfloor \frac{s_k}{2} \right\rfloor + i \cdot s_k, \left\lfloor \frac{s_k}{2} \right\rfloor + j \cdot s_k \right)
\end{equation}
This strategy allows the network to learn from multiple positive samples across different feature scales, which is not only beneficial for objects of varying sizes but also remedies the design limitations of single-positive-sample assignment in classic methods. The head utilizes two parallel branches to predict localization confidence and bounding box parameters, respectively. For the regression, we employ our specifically designed Orientation-Sensitive Loss (OS-Loss), which is tailored for the geo-localization task.

\noindent
\textbf{HBox Geo-localization Head.} To ensure a fair and comprehensive comparison with methods that operate on HBox, we also implement an interchangeable HBox Geo-Localization Head. This auxiliary head adopts a conventional anchor-based framework, where it assigns positive samples based on the IoU between anchor boxes and ground-truth HBoxes. The optimization of the bounding box is guided by a standard IoU-based loss function.

\noindent
\textbf{SAM Prompt Segmentation.} The outputs from either head can be seamlessly integrated with large foundation models. Specifically, the predicted boxes serve as effective prompts for the SAM \cite{sam}. Following the SPS from TROGeo \cite{zhang2025breaking}, we leverage the remarkable zero-shot capability of SAM to achieve precise instance segmentation without requiring any segmentation-specific training or datasets.

\subsubsection{Optimization Objective} 
The overall training objective $\mathcal{L}$ is combination of the regression $\mathcal{L}_{reg}$, classification $\mathcal{L}_{c}$, and center-ness $\mathcal{L}_{cn}$. The formulation is as follows:
\begin{equation}
\begin{split}
\mathcal{L} =& \frac{\mu_1}{N_{pos}} \sum_{l} \mathcal{L}_{c}(c_{l}^{*}, c_{l}) 
 + \frac{\mu_2}{N_{pos}} \sum_{l} \mathcal{L}_{cn}(cn_{l}^{*}, cn_{l}) \\
& + \frac{\mu_3}{\sum cn_{pos}} \sum_{l} \mathds{1}_\mathrm{c_{l} > 0} cn_{l}\mathcal{L}_{reg}\left\{ B^*_l, B_l \right\}
\end{split}
\label{loss-total}
\end{equation}
where \( \mathcal{L}_{c} \) is the focal loss \cite{retina}, and \( \mathcal{L}_{cn} \) is cross-entropy loss. $N_{pos}$ denotes the number of positive samples. The regression loss $\mathcal{L}_{\text{reg}}$ is calculated using the standard IoU loss for the HBox regression. For the RBox regression, we introduce our novel Orientation-Sensitive Loss (OS-Loss).  The hyperparameters $\mu_1$, $\mu_2$, and $\mu_3$ balance the contributions of these terms and are uniformly set to 1.0 in our experiments.

\textbf{Orientation-Sensitive Loss.} To specifically address the challenges in cross-view object geo-localization, we design the Orientation-Sensitive Loss $\mathcal{L}_{o}$ to jointly optimize location and orientation. It is defined as:
\begin{equation}
\mathcal{L}_{o} = \mathcal{L}_{IoU}(B^*_l, B_l) + \alpha \cdot \text{sigmoid}(d_c) 
+ \beta \cdot \left| \sin(\Delta\theta) \right|
\label{osloss}
\end{equation}

The OS-Loss enhances the standard $\mathcal{L}_{\text{IoU}}$ with two critical constraints. The first penalizes the Euclidean distance $d_c$ between the predicted and GT box centers. The second, $|\sin(\Delta\theta)|$, can reflect the actual difference between the predicted box and the GT box, even at the angle-defined boundary. The coefficients $\alpha$ and $\beta$ balance these constraints.

\begin{figure}[t]
\centering
\includegraphics[width=0.35\textwidth]{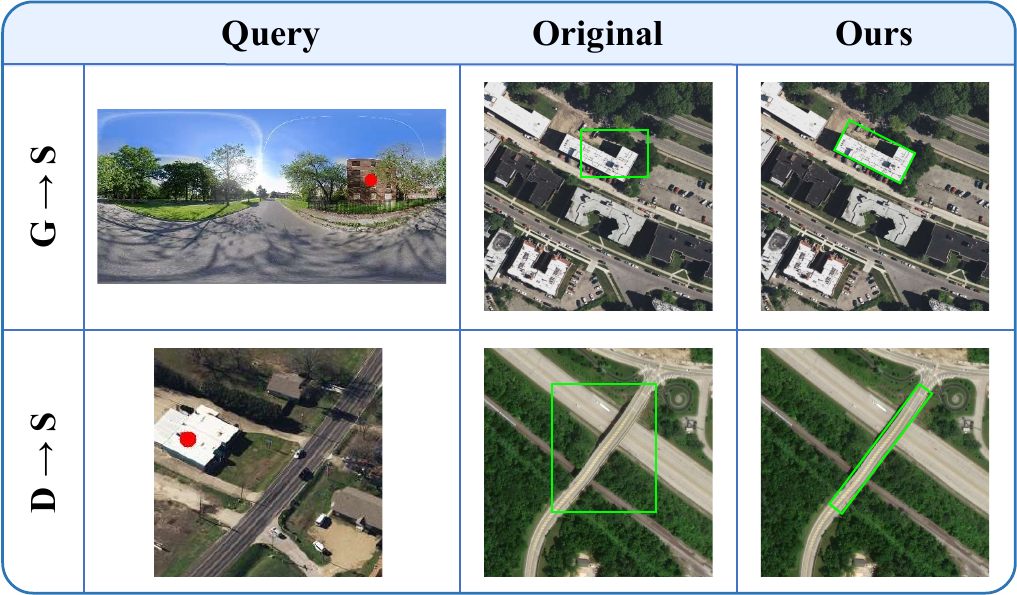}
\caption{Visualization of data annotation for cross-view object geo-localization. Click points are indicated by \textcolor{red}{\emph{red}} dots in the query images. \textcolor{green}{\emph{Green}} boxes indicate different annotation formats.}
\vspace{-3mm}
\label{fig:data}
\end{figure}

\subsection{CVOGL-R Dataset}
\label{sec:cvogl-r}
To support our RBox-based scheme, we also introduce CVOGL-R dataset. The original CVOGL dataset relies on HBoxes, which, as shown in \cref{fig:data}, cannot precisely capture the geometry of oriented objects. 
To address this, we meticulously re-annotated the entire dataset with precise RBoxes. This process not only provides high-fidelity orientation labels but also corrects numerous misalignments in the original annotations. To quantify the cost-effectiveness, we tracked our annotation process: labeling a precise RBox took, on average, 9 seconds per instance. This is an order of magnitude faster than the pixel-level annotation required by segmentation-based TROGeo, and this cost is consistent with the widely accepted claim that “the cost of labeling a pixel-level segmentation annotation is about 15 times larger than labeling a bounding box" \cite{cost2023weakly}. 

\section{Experiment}
\label{sec:ex}

\begin{table*}
\centering
\caption{Performance comparison on the proposed CVOGL-R dataset. OSGeo-R is against retrieval (Ret.), detection (Det.), and segmentation (Seg.) based approaches. OST denotes TROGeo \cite{zhang2025breaking} leveraging segmentation tasks. The $*$ indicates baseline models adapted for RBox prediction. $\spadesuit$ denotes our model equipped with a more powerful backbone. The \textbf{best} and \underline{second-best} results are highlighted.}
\footnotesize
\setlength{\tabcolsep}{1.5mm}
\begin{tabular}{c|l|cc|cc|cc|cc}
\hline
\multirow{3}{*}{} & \multirow{3}{*}{Method} & \multicolumn{4}{c|}{Drone $\rightarrow$ Satellite} & \multicolumn{4}{c}{Ground $\rightarrow$ Satellite} \\
\cline{3-10}
& & \multicolumn{2}{c|}{Validation} & \multicolumn{2}{c|}{Test} & \multicolumn{2}{c|}{Validation} & \multicolumn{2}{c}{Test} \\
\cline{3-10}
& & Acc@50 & Acc@25 & Acc@50 & Acc@25 & Acc@50 & Acc@25 & Acc@50 & Acc@25 \\
\hline
\multirow{4}{*}{\textbf{Ret.}} & L2LTR \cite{jiansuoL2LTR} & 2.00\% & 24.32\% & 1.32\% & 21.84\% &0.51\% & 4.99\% & 0.33\% & 4.76\% \\
& TransGeo \cite{jiansuoTransgeo} & 2.01\% & 21.48\% & 1.45\% & 20.12\% &0.70\% & 5.03\% & 0.40\% & 6.53\% \\
& SAFA \cite{jiansuoSAFA} & 2.39\% & 22.40\% & 1.51\% & 21.47\% &0.71\% & 4.89\% & 0.41\% & 6.51\% \\ 
& Sample4Geo \cite{jiansuosample4geo} & 6.07\% & 32.72\% & 5.34\% & 30.32\% & 0.76\% & 5.93\% & 0.72\% & 6.75\% \\
\hline
\multirow{2}{*}{\textbf{Seg.}} & TROGeo \cite{zhang2025breaking}  &54.32\% & 71.38\% & 56.22\% & 74.21\% & 38.68\% &47.11\% & 37.92\% &  49.31\% \\
& TROGeo + OST \cite{zhang2025breaking} &55.58\% & 72.21\% & 56.53\% & 76.05\% & 40.09\% &50.92\% & 38.75\% &  50.46\% \\
\hline
\multirow{6}{*}{\textbf{Det.}} & DetGeo \cite{jianceDetgeo} & 45.50\% & 57.04\% & 43.47\% & 57.04\% & 35.64\% & 46.26\% & 34.43\% & 44.91\% \\
& DetGeo$^*$ \cite{jianceDetgeo} & 47.35\% & 56.88\% & 50.26\% & 58.99\% & 39.87\% & 45.40\% & 41.83\% & 47.17\% \\
& VAGeo \cite{jianceVAgeo} & 48.65\% & 63.06\% & 49.02\% & 62.59\% & 38.57\% & 47.45\% & 36.79\% &  45.73\% \\
& VAGeo$^*$ \cite{jianceVAgeo} &51.14\% & 61.76\% & 55.50\% & 64.23\% & 40.63\% &48.10\% & 43.27\% &  48.41\% \\
\rowcolor{gray!20}
\cellcolor{white} & OSGeo-R (ours) & \underline{68.36\%} & \underline{72.59\%} & \underline{73.48\%} & \underline{78.42\%} & \underline{48.75\%} & \underline{51.57\%} & \underline{50.36\%} & \underline{52.42\%} \\
\rowcolor{gray!20}
\cellcolor{white} & OSGeo-R $\spadesuit$ (ours) & \textbf{69.01\%} & \textbf{73.56\%} & \textbf{74.31\%} & \textbf{78.62\%} & \textbf{49.40\%} & \textbf{51.79\%} & \textbf{51.39\%} & \textbf{53.34\%} \\
\hline
\end{tabular}
\label{tab:comparison_cvogl-R}
\end{table*}


\begin{table*}
\centering
\caption{Performance comparison on the standard CVOGL dataset, which utilizes HBox annotations.}
\footnotesize
\setlength{\tabcolsep}{1.5mm}
\begin{tabular}{c|l|cc|cc|cc|cc}
\hline
\multirow{3}{*}{} & \multirow{3}{*}{Method} & \multicolumn{4}{c|}{Drone $\rightarrow$ Satellite} & \multicolumn{4}{c}{Ground $\rightarrow$ Satellite} \\
\cline{3-10}
& & \multicolumn{2}{c|}{Validation} & \multicolumn{2}{c|}{Test} & \multicolumn{2}{c|}{Validation} & \multicolumn{2}{c}{Test} \\
\cline{3-10}
& & Acc@50 & Acc@25 & Acc@50 & Acc@25 & Acc@50 & Acc@25 & Acc@50 & Acc@25 \\
\hline
\multirow{4}{*}{\textbf{Ret.}} & L2LTR \cite{jiansuoL2LTR} & 5.96\% & 38.68\% & 6.27\% & 38.95\% & 1.84\% & 12.24\% & 2.16\% & 10.69\% \\
& TransGeo \cite{jiansuoTransgeo} & 5.42\% & 34.78\% & 6.37\% & 35.05\% & 3.25\% & 21.67\% & 2.88\% & 21.17\% \\
& SAFA \cite{jiansuoSAFA} & 6.39\% & 36.19\% & 6.58\% & 37.41\% & 3.25\% & 20.59\% & 3.08\% & 22.20\% \\
& Sample4Geo \cite{jiansuosample4geo} & 16.25\% & 52.87\% & 18.40\% & 52.83\% & 4.33\% & 24.27\% & 5.24\% & 25.80\% \\
\hline
\multirow{2}{*}{\textbf{Seg.}} & TROGeo \cite{zhang2025breaking}  & 65.87\% & \underline{72.38\%} & 68.65\% & 74.31\% & 44.20\% & 48.86\% & 45.53\% & 49.44\% \\
& TROGeo + OST \cite{zhang2025breaking} & \underline{66.63\%} & \textbf{73.35\%} & 68.96\% & \underline{76.16\%} & 46.59\% & \underline{51.46\%} & 46.56\% & 51.08\% \\
\hline
\multirow{5}{*}{\textbf{Det.}} & DetGeo \cite{jianceDetgeo} & 55.15\% & 59.81\% & 57.66\% & 61.97\% & 43.99\% & 46.70\% & 42.24\% & 45.43\% \\
& VAGeo \cite{jianceVAgeo} &59.59\% & 64.25\% & 61.87\% & 66.19\% & 44.42\% &47.56\% & 45.22\% &  48.21\% \\
& HALO \cite{jianceHALO} &63.81\% & 70.21\% & 64.44\% & 71.53\% & 46.91\% &50.81\% & 46.76\% &  \underline{51.59\%} \\
\rowcolor{gray!20}
\cellcolor{white} & OSGeo-H (ours) & 66.52\% & 71.40\% & \underline{69.27\%} & 74.20\% & \underline{47.24\%} & 51.14\% & \underline{47.69\%} & 50.67\% \\
\rowcolor{gray!20}
\cellcolor{white} & OSGeo-H $\spadesuit$ (ours) & \textbf{68.15\%} & 72.26\% & \textbf{72.25\%} & \textbf{76.77\%} &\textbf{49.30\%} & \textbf{52.00\%} & \textbf{48.30\%} & \textbf{51.69\%} \\
\hline
\end{tabular}
\label{tab:comparison_cvogl}
\end{table*}

\subsection{Datasets and Evaluation Metrics}
\textbf{CVOGL.} The CVOGL\cite{jianceDetgeo} is a large-scale cross-view object geo-localization dataset, which uses HBox annotations for query objects. See \textcolor{blue}{\emph{supplementary material}} for details.


\noindent
\textbf{CVOGL-Seg.} The CVOGL-Seg \cite{zhang2025breaking} extends CVOGL dataset by providing instance segmentation masks for all query instances. See \textcolor{blue}{\emph{supplementary material}} for details.


\noindent
\textbf{Evaluation Metrics.} We follow the protocol from Zhang et al. \cite{zhang2025breaking} and use a comprehensive set of metrics to evaluate segmentation quality. See \textcolor{blue}{\emph{supplementary material}} for details.



\subsection{Implementation Details}
Unless otherwise specified, our OSGeo model utilizes a ResNet-50 \cite{resnet50} backbone and is trained with identical hyperparameters. See \textcolor{blue}{\emph{supplementary material}} for details.

\begin{table*}
\centering
\caption{Comparison with previous works on the CVOGL-Seg dataset. ``+ SPS" means that SAM Prompt Stage (SPS) from TROGeo\cite{zhang2025breaking} is added to obtain the segmentation mask using the box output as a prompt. The \textbf{best} and \underline{second-best} results are highlighted.}
\resizebox{\textwidth}{!}{
\begin{tabular}{l|cccc|cccc|cccc|cccc}
\hline
\multirow{4}{*}{Method} & \multicolumn{8}{c|}{Drone $\rightarrow$ Satellite} & \multicolumn{8}{c}{Ground $\rightarrow$ Satellite} \\
\cline{2-17}
& \multicolumn{4}{c|}{Validation} & \multicolumn{4}{c|}{Test} & \multicolumn{4}{c|}{Validation} & \multicolumn{4}{c}{Test} \\
\cline{2-17}
& mIoU $\uparrow$ & mDice $\uparrow$ & AAE $\downarrow$ & ME $\downarrow$ & mIoU $\uparrow$ & mDice $\uparrow$ & AAE $\downarrow$ & ME $\downarrow$ & mIoU $\uparrow$ & mDice $\uparrow$ & AAE $\downarrow$ & ME $\downarrow$ & mIoU $\uparrow$ & mDice $\uparrow$ & AAE $\downarrow$ & ME $\downarrow$ \\
& (\%) & (\%) & ($m^2$) & ($m$) & (\%) & (\%) & ($m^2$) & ($m$) & (\%) & (\%) & ($m^2$) & ($m$) & (\%) & (\%) & ($m^2$) & ($m$) \\
\hline
Sample4Geo \cite{jiansuosample4geo} & 16.83 & 25.41 & 3782.95 & 62.96 & 16.62 & 25.29 & 3910.25 & 57.49 & 5.30 & 8.56 & 3782.69 & 127.92 & 5.92 & 9.56 & 3910.39 & 128.81 \\
Sample4Geo + SPS \cite{jiansuosample4geo} & 19.21 & 25.37 & 2270.92 & 62.51 & 18.67 & 24.83 & 2178.15 & 57.47 & 5.03 & 6.94 & 2403.32 & 128.58 & 6.26 & 8.57 & 2347.28 & 130.22 \\
DetGeo \cite{jianceDetgeo} & 30.81 & 39.40 & 2885.87 & 80.69 & 30.50 & 39.12 & 2517.70 & 85.35 & 25.77 & 32.65 & 2510.07 & 100.89 & 25.40 & 32.27 & 3005.94 & 103.71 \\
DetGeo + SPS \cite{jianceDetgeo} & 42.68 & 48.24 & 1096.89 & 80.19 & 42.27 & 47.98 & 1149.01 & 85.06 & 34.76 & 39.18 & 1229.54 & 100.46 & 34.04 & 38.69 & 1721.25 & 103.39 \\
TROGeo + SPS \cite{zhang2025breaking} & \textbf{55.54} & \textbf{62.80} & \textbf{762.15} & \textbf{49.22} & \underline{56.59} & \underline{64.24} & \underline{927.00} & 50.45 & \underline{38.79} & \underline{43.76} & \underline{1169.61} & \textbf{90.59} & \underline{38.12} & \underline{43.34} & \textbf{1126.56} & \textbf{95.02} \\
\hline
\rowcolor{gray!20}
OSGeo (ours) & 48.98 & 57.67 & 1180.31 & 59.75 & 52.32 & 61.42 & 1116.23 & \underline{48.73} & 35.75 & 41.93 & 1509.71 & 92.33 & 36.20 & 42.44 & 1666.67 & 96.05 \\
\rowcolor{gray!20}
OSGeo + SPS (ours) & \underline{54.46} & \underline{61.18} & \underline{782.74} & \underline{59.42} & \textbf{57.20} & \textbf{64.60} & \textbf{772.24} & \textbf{48.46} & \textbf{39.47} & \textbf{44.26} & \textbf{1097.70} & \underline{92.14} & \textbf{39.45} & \textbf{44.58} & \underline{1290.60} & \underline{95.78} \\
\hline
\end{tabular}}
\label{tab:CVOGL-Seg}
\end{table*}

\subsection{Main Results}
Our OSGeo model has two variants, differing in the head, \textbf{OSGeo-R} for RBox prediction and \textbf{OSGeo-H} for HBox prediction. This dual configuration allows for comprehensive evaluation across datasets with either annotation type.

\noindent
\textbf{Results on CVOGL-R.} We begin by establishing a new benchmark on our proposed CVOGL-R, as detailed in \cref{tab:comparison_cvogl-R}. Retrieval-based methods, including strong performers like Sample4Geo \cite{jiansuosample4geo}, exhibit a significant performance decline. The operational mechanism of these models, relying on matching coarse image grids, proves insufficient for the fine-grained precision required for RBox evaluation. 




Regarding segmentation and detection-based methods, we adopt the evaluation settings from TransGeo \cite{jiansuoTransgeo} to assess their performance on our dataset. TROGeo \cite{zhang2025breaking} shows competitive results, which are further improved by leveraging an auxiliary segmentation task (TROGeo + OST). For detection-based methods like DetGeo \cite{jianceDetgeo}, we adapt them to predict RBoxes, denoted by DetGeo$^*$. The DetGeo \cite{jianceDetgeo} achieves a \textbf{43.47\%} Acc@50 on the Drone→Satellite test set. When the architecture is modified to predict RBoxes, its accuracy rises substantially to \textbf{50.26\%}. 


Our OSGeo-R significantly outperforms all baseline methods across all metrics. On the Drone→Satellite test set, OSGeo-R achieves an accuracy of \textbf{73.48\%} Acc@50, surpassing the adapted DetGeo$^*$ by a remarkable margin of \textbf{23.22\%}. It demonstrates the effectiveness of an architecture specifically designed to perceive object orientation and scale. Furthermore, when equipped with a more powerful backbone (denoted as OSGeo-R $\spadesuit$), the performance of our model is further boosted to \textbf{74.31\%}, setting a new state-of-the-art in cross-view object geo-localization. The consistent and large improvements across both Drone→Satellite and Ground→Satellite subsets affirm the generalization and superiority of our orientation-sensitive design.

\noindent
\textbf{Results on CVOGL.} To make a fair comparison, we evaluate OSGeo-H on the standard CVOGL dataset against existing methods. As presented in \cref{tab:comparison_cvogl}, our OSGeo-H establishes a new state-of-the-art among detection-based approaches, significantly outperforming prior methods like HALO \cite{jianceHALO}. Most notably, OSGeo-H successfully closes the performance gap with the segmentation-based method. On the Drone→Satellite test set, OSGeo-H achieves \textbf{69.27\%} Acc@50, surpassing the leading segmentation method TROGeo + OST \cite{zhang2025breaking} (68.96\%), which relies on costly pixel-level annotations. This result compellingly demonstrates that our framework can deliver state-of-the-art localization precision without the prohibitive expense of mask supervision. Furthermore, our enhanced OSGeo-H $\spadesuit$ decisively outperforms all prior work, including segmentation-based solutions, across almost all metrics.

\begin{figure*}[ht]
\centering
\includegraphics[width=0.7\textwidth]{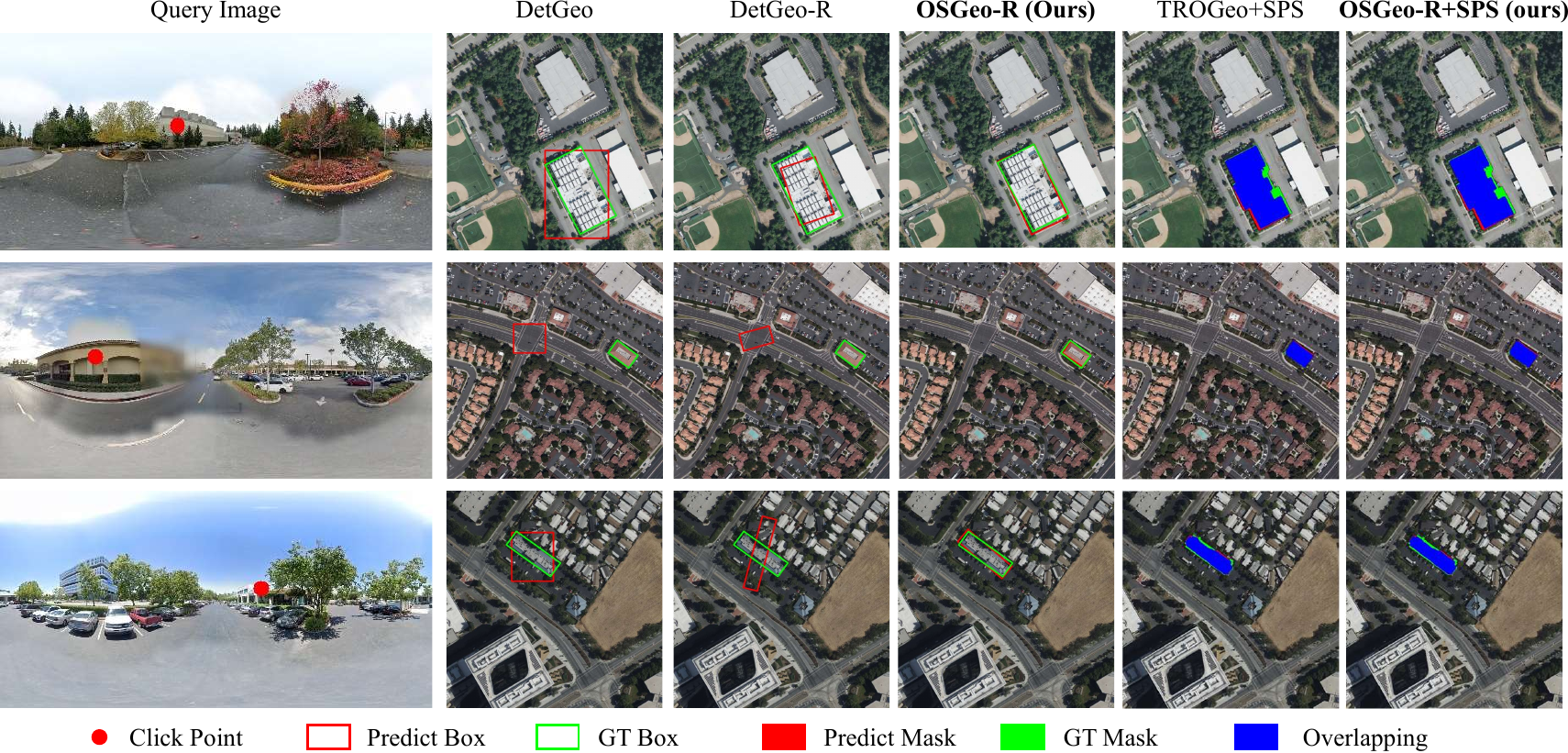} 
\caption{Visual comparison with state-of-the-art methods.}
\label{fig:vis}
\vspace{-2mm}
\end{figure*}

\noindent
\textbf{Results on CVOGL-Seg.} For a fair comparison, we evaluate our model on the CVOGL-Seg dataset. Unlike conventional approaches that rely on expensive pixel-wise masks, our method is trained using only bounding box annotations. To enable segmentation, we integrate the SAM Prompt Stage (SPS) \cite{zhang2025breaking}, which converts our predicted boxes into segmentation masks. The results in \cref{tab:CVOGL-Seg} demonstrate that our OSGeo + SPS rivals the performance of the segmentation-based TROGeo. Specifically, on the Drone→Satellite test set, it surpasses all competitors, achieving a \textbf{57.20\%} mIoU and a \textbf{64.60\%} mDice, while also minimizing AAE of 772.24 $m^2$ and ME of 48.46 $m$. This commanding performance is maintained on the more challenging Ground→Satellite test set, where our model again secures top ranks with 39.45\% mIoU and 44.58\% mDice. These results validate that our approach, despite relying on significantly cheaper box-level supervision, not only rivals but also outperforms methods dependent on masks.

\subsection{Analysis of Cost}
Our experiments compellingly validate our RBox-based paradigm, which resolves the precision-cost dilemma. As quantified in \cref{sec:cvogl-r}, the RBox annotation cost is empirically minimal (avg.9 seconds per object), which we estimate to be an order of magnitude faster than pixel-level masks, a claim consistent with established literature \cite{cost2023weakly}. Crucially, this low cost is not a compromise: \cref{tab:CVOGL-Seg} demonstrates that our OSGeo, trained only on low-cost boxes, surpasses the performance of the segmentation-based TROGeo (trained on expensive masks) on the CVOGL-Seg benchmark (e.g., 57.20\% vs. 56.59\% mIoU on Drone→Sat Test). Furthermore, \cref{tab:comparison_cvogl} reinforces this by showing our OSGeo-H (HBox-trained) also achieves SOTA precision, outperforming TROGeo+OST (e.g., 76.77\% vs 76.16\% Acc@50 on Drone→Sat Test). Collectively, these results prove that our paradigm is a more economical and effective path to high-precision geo-localization.

\subsection{Ablation Studies}
To validate our design choices, we conduct a series of ablation studies on the CVOGL-R test set.

\noindent
\textbf{The Effect of Main Components.} We analyze the individual contributions of our two core innovations, MCP and OS-Loss $\mathcal{L}_{o}$. The results of the ablation study are presented in \cref{tab:components}. We establish a baseline using DetGeo$^*$ that omits all components, which achieves a 50.26\% Acc@50 on the Drone→Satellite set. The integration of the MCP module alone results in a substantial performance gain, elevating the Acc@50 score to \textbf{69.06\%}. This demonstrates the effectiveness of MCP in leveraging multi-scale contextual information. Building upon this, we replace the conventional IoU loss with our proposed $\mathcal{L}_{o}$. This change further boosts the Acc@50 to \textbf{73.48\%}. The consistent improvement across all metrics clearly validates that both MCP and $\mathcal{L}_{o}$ are essential components, and their combination enables the superior performance of our final model. Notably, this advantage becomes magnified with the stricter Acc@75 metric, indicating superior precision in fine-grained localization.

\begin{table}[t]
\centering
\caption{Ablation study on different components of OSGeo. $\mathcal{L}_{o}$ and $\mathcal{L}_{IoU}$ represent the losses in Geo-Localization Decode Head.}
\resizebox{.99\columnwidth}{!}{
\begin{tabular}{ccc|ccc|ccc}
\hline
\multirow{2}{*}{MCP} & \multirow{2}{*}{$\mathcal{L}_{o}$} & \multirow{2}{*}{$\mathcal{L}_{IoU}$} & \multicolumn{3}{c|}{Drone $\to$ Satellite} & \multicolumn{3}{c}{Ground $\to$ Satellite} \\
\cline{4-9}
& & & Acc@75 & Acc@50 & Acc@25 & Acc@75 & Acc@50 & Acc@25 \\
\hline \hline
$\times$    & $\times$    & $\times$    & 10.38\% & 50.26\% & 58.99\% & 8.94\% & 41.83\% & 47.17\% \\
$\times$    & \checkmark  & $\times$    & 31.86\% & 67.42\% & 74.00\% & 22.51\% & 44.09\% & 48.82\% \\
$\times$    & $\times$    & \checkmark  & 29.91\% & 65.16\% & 72.87\% & 19.22\% & 43.17\% & 47.79\% \\
\checkmark  & $\times$    & \checkmark  & 36.07\% & 69.06\% & 75.64\% & 28.67\% & 47.56\% & 50.27\% \\
\checkmark  & \checkmark  & $\times$    & \textbf{42.24\%} & \textbf{73.48\%} & \textbf{78.42\%} & \textbf{29.39\%} & \textbf{50.36\%} & \textbf{52.42\%} \\
\hline
\end{tabular}}
\label{tab:components}
\end{table}

\begin{table}[t]
\centering
\caption{The effect of shared backbone and neck. “Shared." indicates shared parameters, “Param." indicates the parameter count.}
\resizebox{.99\columnwidth}{!}{
\begin{tabular}{lcc|ccc|ccc}
\hline
\multirow{2}{*}{Method} & \multirow{2}{*}{Shared.} & \multirow{2}{*}{Param.(M)$\downarrow$} & \multicolumn{3}{c|}{Drone $\to$ Satellite} & \multicolumn{3}{c}{Ground $\to$ Satellite} \\
\cline{4-9}
& & & Acc@75 & Acc@50 & Acc@25 & Acc@75 & Acc@50 & Acc@25 \\
\hline \hline
DetGeo$^*$ \cite{jianceDetgeo}& $\times$    & 70.38   & 10.38\% & 50.26\% & 58.99\% & 8.94\% & 41.83\% & 47.17\% \\
VAGeo$^*$ \cite{jianceVAgeo} & $\times$  & 70.41   & 9.66\% & 55.50\% & 64.23\% & 9.66\% & 43.27\% & 48.41\% \\
\rowcolor{gray!20}
OSGeo-R (ours)& $\times$    & 56.74  & 38.13\% & 67.83\% & 71.84\% & 26.98\% & 49.19\% & 51.68\% \\
\rowcolor{gray!20}
OSGeo-R (ours)& \checkmark  & \textbf{30.63} & \textbf{42.24\%} & \textbf{73.48\%} & \textbf{78.42\%} & \textbf{29.39\%} & \textbf{50.36\%} & \textbf{52.42\%} \\
\hline
\end{tabular}}
\label{tab:share}
\vspace{-4mm}
\end{table}

\noindent
\textbf{The Effect of Shared Backbone and Neck.} We conduct an ablation study to evaluate the impact of parameter sharing on our model, with the results presented in \cref{tab:share}. The experiment reveals that employing a shared backbone and neck not only drastically reduces the model size but also significantly boosts geo-localization performance. Our shared model consistently achieves higher accuracy across all metrics, improving the Drone→Satellite Acc@75 from 38.13\% to 42.24\% while nearly halving the parameter count from \textbf{56.74M} to \textbf{30.63M}. This performance gain is even more pronounced when compared to traditional methods, such as DetGeo$^*$ and VAGeo$^*$, which do not utilize parameter sharing. The results demonstrate that parameter sharing is not a compromise for efficiency but a direct contributor to performance. By compelling the model to learn a unified representation, it guides the network to focus on the essential viewpoint-invariant features required for cross-view matching, leading to a more effective and efficient model.


\noindent
\textbf{The Effect of Different Backbones.} We conduct an ablation study to assess the influence of backbone architectures on OSGeo-R, evaluating ResNet-18, ResNet-50, and ResNet-101. The results, presented in \cref{tab:backbones}, reveal a direct correlation between network depth and localization accuracy. Specifically, increasing the model capacity from ResNet-18 to ResNet-101 yields substantial performance gains across both test splits. The ResNet-101 based model consistently achieves the best results, securing top scores on both the Drone→Satellite (47.28\% Acc@75) and the more challenging Ground→Satellite (32.27\% Acc@75) evaluations. This confirms our framework effectively capitalizes on the richer feature representations from deeper networks. 

\begin{table}[t]
\centering
\caption{Ablation study of different backbones for OSGeo-R.}
\resizebox{.99\columnwidth}{!}{
\begin{tabular}{l|ccc|ccc}
\hline
\multirow{2}{*}{Backbone} & \multicolumn{3}{c|}{Drone $\to$ Satellite} & \multicolumn{3}{c}{Ground $\to$ Satellite} \\
\cline{2-7}
& Acc@75 & Acc@50 & Acc@25 & Acc@75 & Acc@50 & Acc@25 \\
\hline \hline
ResNet-18    & 33.71\% & 65.57\% & 70.61\% & 24.56\% & 48.20\% & 52.42\% \\
ResNet-50    & 42.24\% & 73.48\% & 78.42\% & 29.39\% & 50.36\% & 52.42\% \\
ResNet-101 & \textbf{47.28\%} & \textbf{74.31\%} & \textbf{78.62\%} & \textbf{32.27\%} & \textbf{51.39\%} & \textbf{53.34\%} \\
\hline
\end{tabular}}
\vspace{-4mm}
\label{tab:backbones}
\end{table}

\subsection{Qualitative Analysis} 

The qualitative results presented in \cref{fig:vis} provide compelling visual evidence for the superior localization capabilities of our framework across objects of varying scales and orientations. DetGeo struggles with oriented objects (Rows 1, 3), producing oversized boxes that include significant background noise and fail to capture the true geometry of the target. In stark contrast, our OSGeo-R generates precise, rotated boxes that tightly conform to the boundaries of the object. Furthermore, OSGeo-R delivers accurate results for both large objects (Row 1) and small, challenging targets where competing methods like DetGeo and DetGeo-R completely fail (Row 2). When extended for segmentation, our OSGeo-R+SPS, \textbf{trained only with bounding boxes}, produces masks of a quality comparable to the segmentation-based TROGeo. See \textcolor{blue}{\emph{supplementary material}} for details.

 
\section{Conclusion}
\label{sec:conclusion}
This work tackles the geometric inadequacy of HBoxes for oriented objects in cross-view object geo-localization. We introduce OSGeo, a novel framework for regressing tight Rotated Bounding Boxes (RBoxes), and release CVOGL-R, the first dataset with precise RBox annotations for CVOGL. Crucially, OSGeo achieves accuracy comparable or superior to costly segmentation methods but with an order-of-magnitude lower annotation cost, demonstrating that explicitly modeling object orientation is a highly effective and economical path to high-precision geo-localization.

{
    \small
    \bibliographystyle{ieeenat_fullname}
    \bibliography{main}
}


\end{document}